\documentclass[conference]{IEEEtran}
\IEEEoverridecommandlockouts
\usepackage{cite}
\usepackage{amsmath,amssymb,amsfonts}
\usepackage{algorithmic}
\usepackage{graphicx}
\usepackage{textcomp}
\usepackage{xcolor}
\usepackage{multirow}

\def\BibTeX{{\rm B\kern-.05em{\sc i\kern-.025em b}\kern-.08em
    T\kern-.1667em\lower.7ex\hbox{E}\kern-.125emX}}
\begin{document}

\title{DAPL: Integration of Positive and Negative Descriptions in Text-Based Person Search}

\author{
    \IEEEauthorblockN{
        Yuchuan Deng\textsuperscript{1}, 
        Zhanpeng Hu\textsuperscript{1}, 
        Zijie Xin\textsuperscript{2}, 
        Chuang Deng\textsuperscript{1}, 
        Qijun Zhao\textsuperscript{1}\thanks{*Corresponding author: Qijun Zhao (qjzhao@scu.edu.cn)}
    }
    \IEEEauthorblockA{
        \textsuperscript{1}Sichuan University, Chengdu, China \\
        \textsuperscript{2}Renmin University of China, Beijing, China \\
        \{dengyuchuan, lucas, dengchuang\}@stu.scu.edu.cn, xinzijie@ruc.edu.cn, qjzhao@scu.edu.cn \\
    }
}

\maketitle

\begin{abstract}
Text-based person search (TBPS) aims to retrieve specific images of individuals from large datasets using textual descriptions. Existing TBPS methods focus primarily on identifying explicit positive attributes, often neglecting the critical role of negative descriptions. This oversight can lead to false positives, where images that should be excluded based on negative descriptions are incorrectly included, due to partial alignment with the positive criteria. To address this limitation, we propose the Dual Attribute Prompt Learning (DAPL) framework, which incorporates both positive and negative descriptions to improve the interpretative accuracy of vision-language models in TBPS tasks. DAPL combines Dual Image-Attribute Contrastive (DIAC) learning with Sensitive Image-Attribute Matching (SIAM) learning to enhance the detection of previously unseen attributes. Furthermore, to achieve a balance between coarse and fine-grained alignment of visual and textual embeddings, we introduce the Dynamic Token-wise Similarity (DTS) loss. This loss function refines the representation of both matching and non-matching descriptions at the token level, providing more precise and adaptable similarity assessments, and ultimately improving the accuracy of the matching process. Empirical results demonstrate that DAPL outperforms state-of-the-art methods, enhancing both precision and robustness in TBPS tasks.
\end{abstract}


\section{Introduction}
\label{sec:intro}
Text-based person search (TBPS)~\cite{li2017person} aims to address the challenge of person re-identification in scenarios where visual data is unavailable, relying solely on textual descriptions for retrieval. 
Recent advances have taken advantage of vision-language models (VLM)~\cite{zhang2024vision} and fine-tuned them with auxiliary tasks focused on pedestrian attributes, resulting in improvements in retrieval accuracy~\cite{jiang2023cross,jiang2023transformer,bai2023rasa,yang2023towards,Qin_2024_CVPR}.

\begin{figure}[htbp]
\centering
\includegraphics[width=\columnwidth]{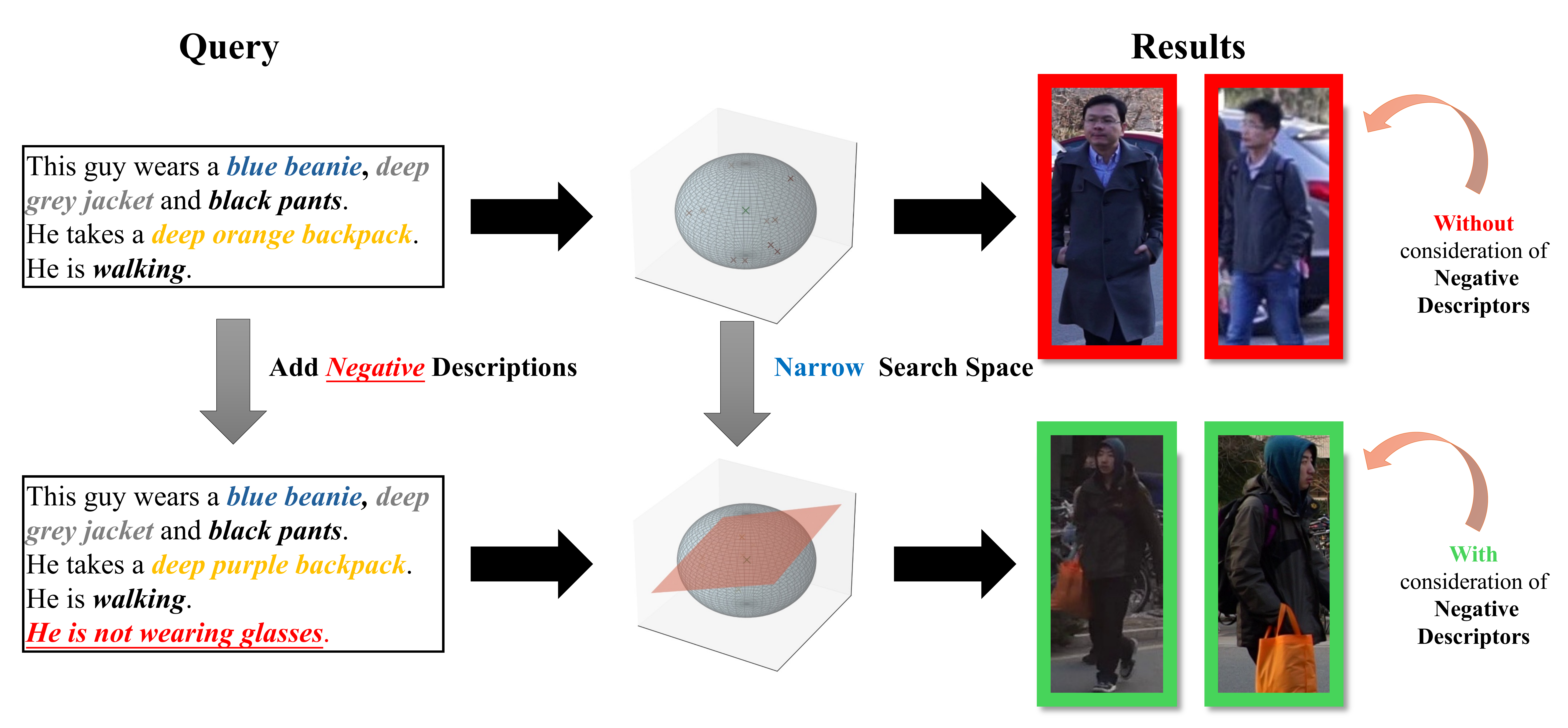}
\caption{Illustration of the impact of negative descriptions: The green box represents a successful retrieval, while the red box indicates a failed retrieval. Negative descriptions provide supplementary information that narrows the search space, leading to more accurate identification of individuals.}
\label{fig:Intro}
\end{figure}

However, as shown in Fig.~\ref{fig:Intro}, traditional TBPS methods primarily focus on identifying explicit positively defined attributes, such as \textit{wearing a gray jacket}. These approaches often neglect the role of negative descriptions, attributes that define the absence of certain features, such as \textit{not wearing glasses}. This neglect can lead to false positives, where images that partly meet the positive criteria are incorrectly included because negative descriptions are not considered adequately.

Introducing negative descriptions adds complexity to the training. While positive descriptions confirm the presence of features (\textit{e.g.}, \textit{wearing a hat}), negative descriptions exclude certain features (\textit{e.g.}, \textit{not carrying a bag}). This dual constraint requires the loss to promote alignment with positive cues while suppressing alignment with negative ones. 
Existing loss functions~\cite{zhang2024vision} treat all attributes equally, neglecting the crucial distinction between positive attributes that should be promoted and negative ones that require suppression.
Furthermore, negative descriptions are typically characterized by ambiguous semantics and strong negation, intended to exclude visually similar images that lack certain key attributes. Nevertheless, current approaches rely on global similarity~\cite{radford2021learning, han2021text, yan2022clip}, which limits their ability to capture the fine-grained cues needed to detect unseen attributes~\cite{yao2021filip}.
Consequently, models struggle to distinguish fine-grained visual and semantic differences, leading to poor performance involving negative descriptions.

To address this challenge, we propose the \textbf{\underline{D}ual \underline{A}ttribute \underline{P}rompt \underline{L}earning (DAPL)} framework. 
DAPL comprises two core components: Dual Image-Attribute Contrastive (DIAC) learning and Sensitive Image-Attribute Matching (SIAM) learning. DIAC improves the detection of unseen attributes, especially those from negative descriptions, while SIAM focuses on key features, reducing the search space and mitigating the impact of negative cues to retain correct candidates.
Additionally, we propose a novel token-level similarity function, Dynamic Token-wise Similarity (DTS) loss, for fine-grained alignment between visual and textual representations.
Our contributions can be summarized as follows:

\textbf{(1)} We introduce negative descriptions into the TBPS matching process to address the issue of excluding inappropriate candidate images. This innovation is implemented in the DAPL framework, comprising the DIAC and SIAM components.

\textbf{(2)} We propose the DTS loss, a novel token-wise similarity function that enables fine-grained alignment between visual and textual representations. 
This improves text-image matching at the attribute level, especially in detecting unseen attributes and narrowing the search space.

\textbf{(3)} Experimental results demonstrate that DAPL enhances TBPS precision by integrating both positive and negative descriptions, leading to significant improvements in matching accuracy and robustness.

\section{Related work}
\subsection{Vision-Language Pretrained Models}
Vision-Language Pretrained Models~\cite{zhang2024vision} have leveraged large-scale image-text pair datasets to explore the complex semantic interactions between visual and textual modalities. In this paradigm, models undergo training across various tasks, including image-text contrastive learning~\cite{radford2021learning} and masked language modeling~\cite{devlin2018bert}, resulting in the generation of representations that are both contextually nuanced and semantically rich. These representations are crucial for improving the performance of text-based person retrieval tasks~\cite{han2021text}.

\subsection{Text-Based Person Search}
Text-based person search (TBPS)~\cite{li2017person} requires distinguishing between numerous pedestrians with subtle inter-class differences while addressing the modality gap between images and textual descriptions. To tackle these challenges, researchers have focused on two main directions: semantic alignment and noise mitigation.
For semantic alignment, methods such as CLIP-based alignment~\cite{yan2022clip} and cross-modal matching~\cite{jiang2023cross,li2023dcel} leverage attention mechanisms to align fine-grained semantics between modalities. For noise mitigation, optimized objective techniques~\cite{bai2023rasa, wang2024fine, Qin_2024_CVPR, park2025plot} use targeted loss functions to bridge modality gaps and enhance feature discrimination. Additionally, generative approaches~\cite{yang2023towards} improve robustness by generating synthetic captions to augment the training process.
However, existing methods focus mainly on explicit feature alignment, neglecting negative matching. To address this, we propose the Dual Attribute Prompt Learning (DAPL) strategy, which integrates both positive and negative descriptions for more comprehensive cross-modal alignment and matching.

\begin{figure*}[htbp]
\centering
\includegraphics[width=\textwidth]{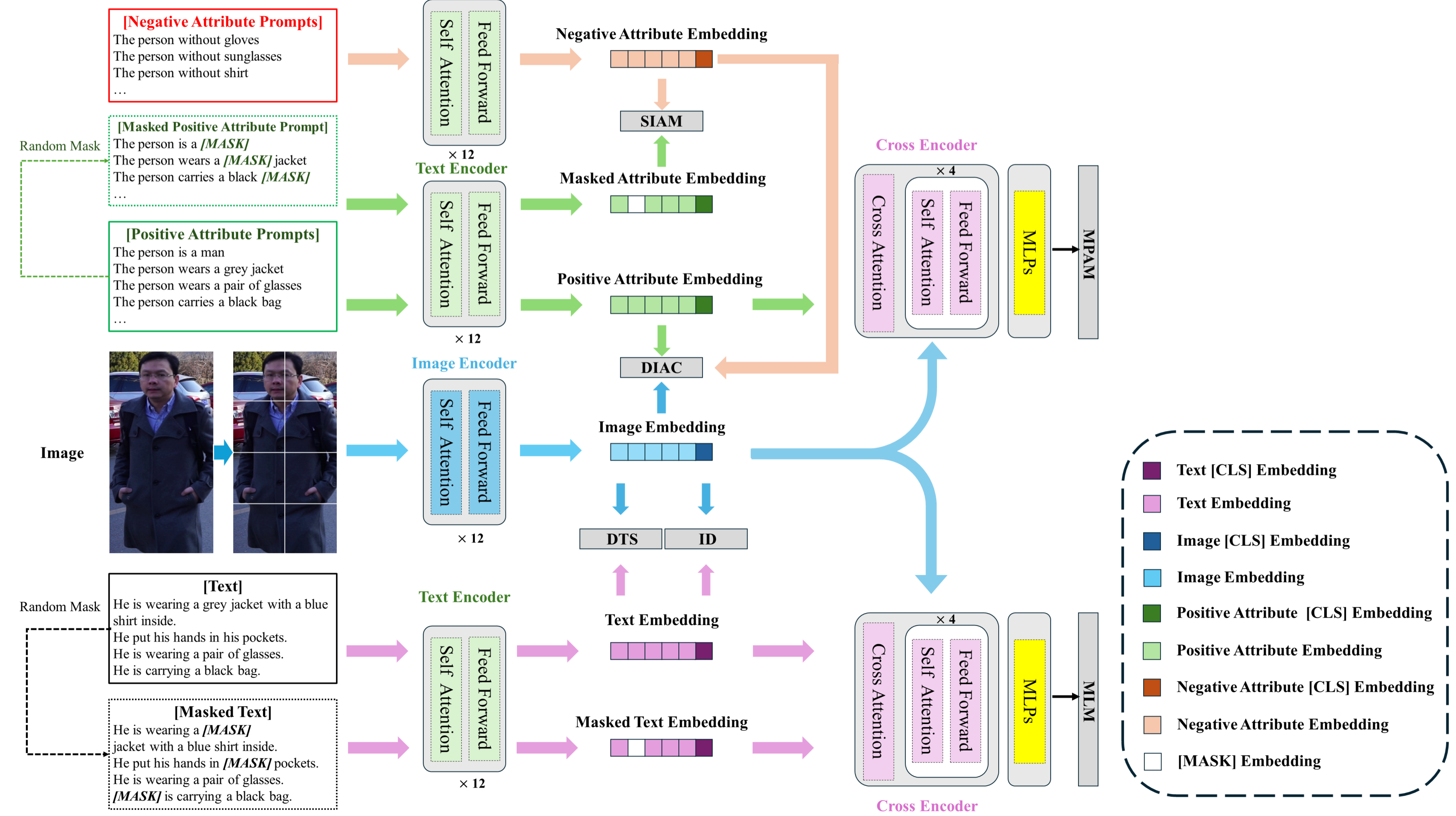}
\caption{Overview of our proposed Dual Attribute Prompt Learning (DAPL) framework. The framework consists of six encoders: one Image Encoder ($E_I$), three Text Encoders ($E_T$), and two Cross Encoders ($E_C$). The training strategies include: Dual Image-Attribute Contrastive Learning (DIAC) for detecting unseen attributes; Sensitive Image-Attribute Matching Learning (SIAM) for filtering out visually similar but semantically mismatched image candidates; Dynamic Token-wise Similarity Loss (DTS) for refining token-level similarity calculations; Masked Positive Attribute Language Modeling (MPAM) and Masked Language Modeling (MLM) for enhancing language understanding by predicting missing words; and Identity Loss (ID) for preserving individual characteristics across modalities. During inference, the framework leverages only the text-image streams, utilizing the pre-trained Image Encoder and Text Encoders to compute the similarity between the text embeddings and the visual embeddings of all images, ranking the candidates based on the similarities.}
\label{fig:framework}
\end{figure*}

\section{Method}
\subsection{Negative Description Generation.} 
For convenience, we define the training dataset $\mathcal{D} = \left\{ \left( \mathbf{I}_i, \mathbf{T}_i \right) \right\}_{i=1}^{N}$ consists of $N$ image-text pairs, where $\mathbf{I}_i$ denotes an image of a person, and $\mathbf{T}_i$ is its corresponding textual description.  
We propose a simple pipeline for generating negative descriptions, using a predefined attribute template $\mathcal{A}$.  
We first analyze descriptions from the datasets to identify verb-noun pairs that represent common attributes describing people, excluding specific details such as colors and constructing $\mathcal{A}$, a predefined attribute template containing $38$ common attributes (\textit{e.g.}, \textit{wearing a hat}, \textit{carrying a backpack})~\cite{yang2023towards}.  

Given a textual description $\mathbf{T}_i$, positive attributes $\mathbf{A}^p_i$ are extracted by matching phrases such as \textit{wearing a white shirt} to their general forms in $\mathcal{A}$ (\textit{e.g.}, \textit{wearing a shirt}). This mapping ensures semantic alignment while avoiding overfitting to specific details like color. Attributes in $\mathcal{A}$ that are not present in $\mathbf{T}_i$ are denoted as negative attributes $\mathbf{A}^n_i$ by appending negation terms, representing absent traits (\textit{e.g}., \textit{without shirt}, \textit{without backpack}). 
Once positive and negative attributes are identified, negative descriptions are generated using the following steps:  
1) Randomly select two distinct negative attributes from $\mathbf{A}^n_i$ to form a description.  
2) Repeat this process three times for each image to create unique, non-overlapping negative descriptions.  
3) Pair each negative description with its corresponding image during training.  
For example, if $\mathbf{T}_i$ is \textit{a person wearing a red shirt}, the positive attribute \textit{wearing a red shirt} is mapped to the template attribute \textit{wearing a shirt}. Based on $\mathcal{A}$, possible negative descriptions may include: \textit{without hat and without backpack} and \textit{without sunglasses and without scarf}.

\subsection{Overview of the Methodology}
The proposed DAPL framework (Fig.~\ref{fig:framework}) consists of six encoders: one Image Encoder ($E_I$), three weight-shared Text Encoders ($E_T$), and two weight-shared Cross Encoders ($E_C$). Each input image $\mathbf{I}_i$ is divided into patches and processed by $E_I$, with a learnable token $\mathbf{v}_{cls}^i$ capturing global features. The resulting visual representation is:
\begin{equation}
\mathbf{F}_{I}^i = \{\mathbf{v}_{cls}^i, \mathbf{v}_1^i, \cdots, \mathbf{v}_{n_v}^i\}.
\end{equation}
For the textual description $\mathbf{T}_i$, the resulting representation includes $[SOS]$ and $[EOS]$ tokens:
\begin{equation}
\mathbf{F}_{T}^i = \{\mathbf{t}_{sos}^i, \mathbf{t}_1^i, \cdots, \mathbf{t}_{n_t}^i, \mathbf{t}_{eos}^i\}.
\end{equation}
Positive attribute prompts $\mathbf{F}_{A^p}^i$ encode explicit attributes, while negative attribute prompts $\mathbf{F}_{A^n}^i$ capture attributes from generated negative descriptions. During training, key strategies are employed: DIAC encourages detection of unseen attributes, especially negatives; SIAM ensures proper weighting of critical features; DTS refines token-level alignment; MPAM and MLM enhance language understanding; and Identity Loss (ID) preserves individual characteristics across modalities. In inference, only the text-image streams are used.

\subsection{Dynamic Token-wise Similarity (DTS) Loss}  
Two individuals may share highly similar positive attributes, such as \textit{wearing a shirt} and \textit{carrying a bag}, but differ in a single negative attribute: one \textit{is without hat}, while another \textit{is wearing a hat}. 
Traditional coarse-grained training methods, such as CLIP~\cite{radford2021learning}, often fail to capture these subtle distinctions during fine-grained attribute alignment~\cite{zhang2024vision}. This limitation is further amplified when negative descriptions—attributes specifying the absence of certain features—are introduced, as they add another layer of complexity to the alignment process.

To address this, token-level similarity computations~\cite{yao2021filip} provide the granularity needed to independently evaluate and prioritize each attribute, whether positive or negative, during alignment. Building on this, we extend the SDM loss~\cite{jiang2023cross} and propose the \textit{Dynamic Token-wise Similarity (DTS) Loss}. By operating at the token level, the DTS Loss introduces fine-grained control, emphasizing the most discriminative attributes and seamlessly integrating both positive and negative constraints. This design ensures the model dynamically adapts to the inherent asymmetry between text-to-image and image-to-text alignment, enabling robust cross-modal matching.

To capture fine-grained cross-modal relationships, we compute token-level similarities given a batch of image and text features $\{\left(\mathbf{F}_{I}^i, \mathbf{F}_{T}^j\right), y_{i,j}\}_{j=1}^B$, where $y_{i,j}$ indicates whether the image-text pair matches ($1$ for a match, $0$ otherwise). For each image token $\mathbf{v}_k^i$, its maximum similarity to all text tokens $\mathbf{t}_r^j$ is defined as:
\begin{equation}
\label{eq:image_to_text_similarity}
\xi_{i,j}^I = \frac{1}{n_v^i}\sum_{k=1}^{n_v^i} \max_{0 \leq r < n_t^j} \mathrm{sim}(\mathbf{v}_k^i, \mathbf{t}_r^j),
\end{equation}
where $\mathrm{sim}(\cdot)$ denotes cosine similarity. This formulation ensures that the most discriminative visual tokens—whether corresponding to positive or negative attributes—are prioritized during alignment.
To calculate the image-to-text matching probability $p_{i,j}^{i2t}$, a softmax function is applied over the computed similarities:
\begin{equation}
\label{eq:probability_image_to_text}
p_{i,j}^{i2t} = \frac{\exp \left(\xi_{i,j}^I / \tau \right)}{\sum_{k=1}^B \exp \left(\xi_{i,k}^I / \tau \right)},
\end{equation}
where $\tau$ is a temperature parameter controlling the sharpness of the probability distribution. Similarly, the text-to-image matching probability can be computed by swapping image and text features. The image-to-text loss is then defined as:
\begin{equation}
\label{eq:dts_loss_i2t}
\mathcal{L}_{i2t} = \frac{1}{B} \sum_{i=1}^B \sum_{j=1}^B p_{i,j}^{i2t} \log \left( \frac{p_{i,j}^{i2t}}{q_{i,j}^{i2t} + \epsilon} \right),
\end{equation}
where $q_{i,j}^{i2t} = y_{i,j} / \sum_{k=1}^B y_{i,k}$ represents the true matching probability, and $\epsilon$ is a constant to ensure numerical stability. Similarly, the text-to-image loss $\mathcal{L}_{t2i}$ is defined in the same manner. The final DTS Loss combines these two components:
\begin{equation}
\label{eq:bidirectional_dts_loss}
\mathcal{L}_{dts} = \mathcal{L}_{i2t} + \mathcal{L}_{t2i}.
\end{equation}
The DTS loss integrates positive and negative attributes for unified token-level alignment, dynamically balancing text-to-image and image-to-text matching. Negative descriptions constrain visually similar but semantically mismatched candidates, while positive attributes enhance the model’s ability to identify key identity-defining features.

\subsection{Dual Attribute Prompt Learning (DAPL)}  
The proposed Dual Attribute Prompt Learning (DAPL) framework addresses the limitations of existing TBPS methods by explicitly incorporating both positive and negative attribute descriptions. Unlike prior works that focus solely on positive attributes or global alignment, DAPL achieves robust attribute-level alignment with minimal computational overhead by introducing two novel components: \textit{Dual Image-Attribute Contrastive (DIAC)} learning and \textit{Sensitive Image-Attribute Matching (SIAM)} learning.

\textbf{Dual Image-Attribute Contrastive (DIAC) Learning.}  
Traditional contrastive learning methods focus primarily on aligning present (positive) attributes, often neglecting absent (negative) attributes that are essential for distinguishing visually similar but semantically distinct samples. DIAC addresses this issue by introducing a balanced contrastive learning objective that aligns visual features with positive attributes while explicitly suppressing negative attributes. This push-pull mechanism ensures fine-grained alignment without significantly increasing computational complexity.

Given an image $\mathbf{I}_i$, token-level similarities are computed between its visual representation $\mathbf{F}_I^i$ and both positive ($\mathbf{F}_{A_p}^i$) and negative ($\mathbf{F}_{A_n}^i$) attribute embeddings:
\begin{equation}
S^{i2a}_{p} = p^{i2t}(\mathbf{F}_{I}^i, \mathbf{F}_{A_p}^i), \quad S^{i2a}_{n} = p^{i2t}(\mathbf{F}_{I}^i, \mathbf{F}_{A_n}^i).
\end{equation}

The contrastive loss for positive and negative attributes is defined as:
\begin{equation}
\mathcal{L}_{piac} = -\frac{1}{2B} \sum_{A_i^p \in \mathcal{A}} \log S^{i2a}_p, \quad
\mathcal{L}_{niac} = -\frac{1}{2B} \sum_{A_i^n \in \mathcal{A}} \log S^{i2a}_n. 
\end{equation}

The final DIAC loss combines these terms:
\begin{equation}
\mathcal{L}_{diac} = \frac{1}{2} (\mathcal{L}_{piac} - \mathcal{L}_{niac}).
\end{equation}
DIAC symmetrically incorporates positive and negative attributes but handles them asymmetrically, suppressing absent features to ensure robustness against visually similar, semantically mismatched candidates.

\textbf{Sensitive Image-Attribute Matching (SIAM) Learning.}  
Textual ambiguity often arises from the dominance of frequent attributes during training, leading to a bias that overlooks rare but critical features. SIAM addresses this by introducing a dynamic weighting mechanism that balances the importance of positive and negative attributes based on their relative frequency in the dataset.

The dynamic adjustment factor $\gamma_a^i$ is defined as:
\begin{equation}
\gamma_a^i = \frac{\mathrm{Count}(A_p^i)}{\mathrm{Count}(A_n^i)},
\end{equation}
where $\mathrm{Count}(A_p^i)$ and $\mathrm{Count}(A_n^i)$ represent the occurrence frequencies of positive and negative attributes, respectively.

Attribute probabilities $\mathbf{p}^a_{i,j}$ are computed as:
\begin{equation}
\mathbf{p}^a_{i,j} = \frac{1}{2} \sum_{k \in \{p, n\}} \mathrm{softmax} \left( \gamma_a^i S^{i2a}_k - \frac{1}{\gamma_a^i} S^{a2i}_k \right),
\end{equation}
where $S^{i2a}_k$ and $S^{a2i}_k$ represent the image-to-attribute and attribute-to-image similarities, respectively.

The SIAM loss is defined as:
\begin{equation}
\mathcal{L}_{siam} = -\frac{1}{B} \sum_{\mathbf{A}_i \in \mathcal{A}} \sum_{j=1}^B \left( y_{i,j} \log(\mathbf{p}^a_{i,j}) + (1 - y_{i,j}) \log(1 - \mathbf{p}^a_{i,j}) \right),
\end{equation}
where $y_{i,j}$ indicates whether the attribute matches the corresponding image.
SIAM dynamically adjusts attribute importance, ensuring subtle but critical features are not overshadowed by frequent attributes, thereby enhancing the model’s sensitivity to fine-grained textual descriptions and its ability to resolve overlapping semantics.

\textbf{Masked Positive Attribute Language Modeling.}
Masked Positive Attribute Modeling (MPAM) refines attribute predictions by focusing on masked positive attributes, using implicit relationship reasoning loss~\cite{li2023dcel} for improved semantic inference. This focus on positive attributes is due to their higher discriminative value, as they directly indicate key traits (\textit{e.g.}, \textit{red shirt}) critical for accurate matching. However, negative attributes (\textit{e.g.}, \textit{no hat}) primarily exclude irrelevant candidates and offer less semantic complexity. Including negative attributes in language modeling provides marginal gains but increases model complexity.
The MPAM loss is defined as: 
\begin{equation} \mathcal{L}_{mpam} = - \sum_{i \in \hat{\mathcal{A}}^p} \log P(\hat{\mathbf{t}}_i | \mathbf{F}_{A^p}^i, \mathbf{F}_I)
\end{equation}
where $\hat{\mathbf{t}}_i$ denotes the masked positive attribute, and $P(\hat{\mathbf{t}}_i | \mathbf{F}_{A^p}^i, \mathbf{F}_I)$ is the conditional probability of predicting the masked positive attribute given the positive attribute representation $\mathbf{F}_{A^p}^i$ and the image embedding $\mathbf{F}_I$.

The combined DAPL loss integrates the three components:
\begin{equation}
\mathcal{L}_{dapl} = \frac{1}{3} (\mathcal{L}_{diac} + \mathcal{L}_{siam} + \mathcal{L}_{mpam}).
\end{equation}

\subsection{Overall Loss Function.} The training loss combines DAPL with other objectives:
\begin{equation}
\mathcal{L} = \lambda_{dts} \mathcal{L}_{dts} + \lambda_{mlm} \mathcal{L}_{mlm} + \lambda_{id} \mathcal{L}_{id} + \lambda_{dapl} \mathcal{L}_{dapl},
\end{equation}
where $\mathcal{L}_{id}$ represents ID loss~\cite{zheng2020dual}, and $\mathcal{L}_{mlm}$ denotes masked language modeling loss~\cite{devlin2018bert}. Hyperparameters $\lambda_{dts}$, $\lambda_{mlm}$, $\lambda_{id}$, and $\lambda_{dapl}$ control the contribution of each component.

\begin{table*}[htbp]
\centering  
\caption{Performance (\%) comparisons on the CUHK-PEDES, ICFG-PEDES, and RSTPReid. \textbf{Bold} and \underline{underline} denote the best and the second best. "-"indicates that the original paper did not use that specific metric to evaluate its models.}
\label{tab_compare}
\begin{tabular}{|l|l|cccc|cccc|cccc|}
\hline
\multirow{2}{*}{\textbf{Methods}} &\multirow{2}{*}{\textbf{Ref}} &\multicolumn{4}{c|}{\textbf{CUHK-PEDES}} &\multicolumn{4}{c|}{\textbf{ICFG-PEDES}} &\multicolumn{4}{c|}{\textbf{RSTPReid}}\\
                            &         &Rank-1&Rank-5 &Rank-10 &mAP &Rank-1 &Rank-5 &Rank-10 &mAP &Rank-1 &Rank-5 &Rank-10 &mAP\\
\hline
CFine~\cite{yan2022clip}    &TIP'23   &69.57 &85.93 &91.15 &-                          &60.83 &76.55 &82.42 &-     &50.55 &72.50 &81.60 &-  \\
IRRA~\cite{jiang2023cross}  &CVPR'23  &73.38 &89.93 &93.71 &66.13                      &63.46 &80.25 &85.82 &38.06 &60.20 &81.30 &88.20 &47.17 \\
RaSa~\cite{bai2023rasa}     &IJCAI'23 &76.51 &90.29 &\underline{94.25} &\textbf{69.38} &65.28 &80.40 &85.12 &\textbf{41.29} &66.90 &\underline{86.50} &91.35 &52.31\\
DECL~\cite{li2023dcel}      &MM'23    &75.02 &\textbf{90.89} &\textbf{94.52} &-        &64.88 &81.34 &86.72 &-     &61.35 &83.95 &90.45 &-  \\
APTM~\cite{yang2023towards} &MM'23    &\underline{76.53} &90.04 &94.15 &66.91          &\textbf{68.51 }&\underline{82.09} &\textbf{87.56} &\underline{41.22} &\underline{67.50} &85.70 &\underline{91.45} &\textbf{52.56}\\
RDE~\cite{Qin_2024_CVPR}    &CVPR'24  &75.94 &90.14 &94.12 &67.56                      &67.68 &\textbf{82.47} &\underline{87.36} &40.06 &65.35 &83.95 &89.90 &50.88\\
FSRL~\cite{wang2024fine}    &ICMR'24  &74.86 &89.97 &94.14 &67.57                      &64.93 &80.71 &86.19 &40.67 &60.65 &83.05 &89.60 &48.18\\
PLOT~\cite{park2025plot}    &ECCV'24  &75.28 &90.42 &94.12 &-                          &65.76 &81.39 &86.73 &-     &61.80 &82.85 &89.45 &- \\
\hline
\textbf{DAPL}               &-        &\textbf{77.43} &\underline{90.73} &94.20 &\underline{68.35} &\underline{67.87} &81.93 &87.13 &40.13 &\textbf{69.12} &\textbf{86.68} &\textbf{92.31} &\underline{52.53} \\
\hline
\end{tabular}
\label{tab:Comparison}
\end{table*}

\begin{table*}[htbp]
\centering
\caption{Ablation experimental results (\%) on the effectiveness of each component in DAPL.} 
\begin{tabular}{|c|l|ccc|ccc|ccc|ccc|}
\hline  
\multirow{2}{*}{No.} & \multirow{2}{*}{Methods} & \multicolumn{3}{c|}{Components} & \multicolumn{3}{c|}{CUHK-PEDES} & \multicolumn{3}{c|}{ICFG-PEDES} & \multicolumn{3}{c|}{RSTPReid} \\
& & $\mathcal{L}_{dts}$  &$\mathcal{L}_{diac}$ &$\mathcal{L}_{siam}$ &Rank-1 &Rank-5 & Rank-&Rank-1 &Rank-5 & Rank-10 &Rank-1 &Rank-5 & Rank-10 \\ 
\hline
1 & Baseline                  &           &           &                          & 73.38         & \underline{89.93} & 93.71            & 63.46         & 80.25            & 85.82           & 60.20           & 81.30            & 88.20\\ 
2 & +$\mathcal{L}_{dts}$      &\checkmark &           &                          & 74.09         & 90.85             &\underline{93.89} & 65.24         & 80.63            & \underline{86.01} & 65.73         & 75.32            & 87.28\\
3 & +$\mathcal{L}_{diac}$     &           &\checkmark &                          & 73.77         & 89.72             & 92.98            & 62.56         & 79.35            & 84.71           & 59.24           & 82.26            & 87.55\\
4 & +$\mathcal{L}_{siam}$     &           &           &\checkmark                & 74.13         & 88.81             & 93.31            & 64.92         & 81.04            & 85.18           & 68.27           & 82.90            & 86.59\\
5 & +$\mathcal{L}_{dts}$ + $\mathcal{L}_{siam}$ &\checkmark  & &\checkmark       &74.65          & 88.37          & 93.62            & 64.49         &\underline{81.38} & 85.04           &\underline{68.62}& 81.75            & 87.30\\
6 & +$\mathcal{L}_{dapl}$                     &           &\checkmark &\checkmark&\underline{75.20} & 89.32             & 93.81            &\underline{66.76} & 81.31         & 85.68           & 67.07           &\underline{83.11} & \underline{89.98}\\
\hline
7 & \textbf{DAPL}        &\checkmark &\checkmark &\checkmark                     &\textbf{77.43} &\textbf{90.73}     &\textbf{94.20}    &\textbf{67.87} &\textbf{82.06}    &\textbf{87.33}   &\textbf{69.12}   &\textbf{86.68}    & \textbf{92.31}\\ 
\hline
\end{tabular}
\label{tab:ablation}
\end{table*}
\section{Experiments}
\subsection{Experimental Setup}
\textbf{Datasets and Evaluation Metrics.} We evaluate our method on three datasets: CUHK-PEDES~\cite{li2017person}, ICFG-PEDES~\cite{ding2021semantically}, and RSTPReid~\cite{zhu2021dssl}.
The primary metric is Rank-$k$, which measures the likelihood of a correct match in the top-$k$ results of a text query. Additionally, we use mean average precision (mAP) for a more comprehensive evaluation.

\textbf{Implementation Details.} We use the CLIP-ViT-B/16 model as the pre-trained image encoder and the CLIP text transformer as the text encoder~\cite{radford2021learning}, supplemented by a cross encoder~\cite{jiang2023cross,li2023dcel}. 
Following~\cite{jiang2023cross,li2023dcel}, input images are resized to $384 \times 128$ and augmented with random horizontal flipping, padding-based cropping, random erasing, and normalization. The maximum token sequence length $n_t$ is set to $77$.
For optimization, we employ the Adam optimizer over $50$ epochs, starting with a learning rate of $1 \times 10^{-5}$ and using cosine decay for gradual adjustment. The first $5$ epochs serve as a warm-up, linearly increasing the learning rate from $1 \times 10^{-6}$ to $1 \times 10^{-5}$. For modules without pre-trained weights, the learning rate is set to $5 \times 10^{-5}$. In the DTS loss calculation, the temperature parameter $\tau$ is set to $0.02$. Regularization parameters $\lambda_{dts}$ and $\lambda_{dapl}$ are set to $2$ and $0.8$, respectively, while $\lambda_{mlm}$ and $\lambda_{id}$ are both set to $1$. 
The model is implemented in PyTorch and trained on 4 NVIDIA RTX 4090 GPUs (24GB each).

\subsection{Results Analysis}
To evaluate DAPL, we compare it with methods that use pre-trained vision-and-language large models as their backbone. Table~\ref{tab:Comparison} summarizes the results.
Specifically, to ensure a fair comparison, we perform inference using only the original positive descriptions, excluding any negative descriptions, in this experiment.
\textbf{CUHK-PEDES:} DAPL achieves the highest Rank-1 accuracy of 77.43\% and a competitive mAP of 68.35\%, outperforming all methods, including APTM~\cite{yang2023towards}. This improvement is attributed to DAPL's effective filtering of visually similar but semantically mismatched candidate images. Unlike APTM, DAPL does not rely on pre-training with the MALS dataset, highlighting its efficiency and adaptability.
\textbf{ICFG-PEDES:} DAPL achieves a Rank-1 accuracy of 67.87\%, slightly below APTM’s 68.51\%, due to face anonymization in the dataset, which limits feature extraction. Despite this, DAPL remains competitive, showcasing its potential for further improvement in datasets with obscured features.
\textbf{RSTPReid:} On the RSTPReid dataset, DAPL achieves a Rank-1 accuracy of 69.12\%, surpassing APTM. It also records high Rank-5 and Rank-10 accuracy (86.68\% and 92.31\%, respectively) and a competitive mAP of 52.53\%, demonstrating robustness in challenging scenarios.
Overall, by incorporating negative descriptions, DAPL enhances fine-grained attribute matching without extensive pre-training, demonstrating its generalizability and effectiveness for TBPS tasks.

\subsection{Ablation Study}
\begin{table*}[htbp]
\centering
\caption{Performance (\%) comparisons of methods with negative descriptions on the CUHK-PEDES, ICFG-PEDES, and RSTPReid datasets.}
\label{tab_negative}
\begin{tabular}{|l|cccc|cccc|cccc|}
\hline
\multirow{2}{*}{Methods} & \multicolumn{4}{c|}{CUHK-PEDES} & \multicolumn{4}{c|}{ICFG-PEDES} & \multicolumn{4}{c|}{RSTPReid} \\
& Rank-1 & Rank-5 & Rank-10 & mAP & Rank-1 & Rank-5 & Rank-10 & mAP & Rank-1 & Rank-5 & Rank-10 & mAP \\
\hline
IRRA~\cite{jiang2023cross}  & 76.82 & 91.23 & 95.58 & 65.89 & 67.04 & 82.28 & 87.66 & 39.83 & 65.88 & 82.29 & 93.88 & 50.28 \\
DECL~\cite{li2023dcel}      & 77.06 & \underline{92.37} & 96.16 & 68.43 & 68.73 & 83.33 & 88.58 & 39.42 & 64.16 & 85.45 & 92.01 & 53.81 \\
RaSa~\cite{bai2023rasa}     & 78.64 & 92.33 & 96.67 & 69.28 & 69.27 &\underline{84.42} & 89.44 &\underline{42.91} & 67.17 & 86.45 & 91.19 & 54.97 \\
APTM~\cite{yang2023towards} &\underline{79.53} & 92.19 & \underline{96.91} & \underline{70.02} & 70.42 & 84.35 & \underline{89.59} & 42.37 &\underline{69.07} & \underline{86.47} & \underline{94.73}  &\underline{55.20}\\
\hline
\textbf{DAPL}               & \textbf{80.05} & \textbf{92.65} & \textbf{97.48} & \textbf{70.24} & \textbf{71.18}  & \textbf{84.68} & \textbf{91.22}  & \textbf{42.96} & \textbf{70.84} & \textbf{86.96} & \textbf{95.32} & \textbf{55.98} \\
\hline
\end{tabular}
\end{table*}

To evaluate the effectiveness of each component in the DAPL framework, we conduct an ablation study on the CUHK-PEDES, ICFG-PEDES, and RSTPReid datasets, summarized in Table~\ref{tab:ablation}. The baseline model is IRRA~\cite{jiang2023cross}.

\textbf{Effectiveness of Dynamic Token-wise Similarity (DTS):}  
Adding the DTS loss to the baseline improves Rank-1 accuracy across all datasets, achieving 74.09\% on CUHK-PEDES, 65.24\% on ICFG-PEDES, and 65.73\% on RSTPReid. DTS dynamically prioritizes token-level similarities, enhancing fine-grained alignment and robustness in challenging scenarios.

\textbf{Impact of Dual Attribute Prompt Learning (DAPL):}  
The DAPL framework, integrating \(\mathcal{L}_{diac}\) and \(\mathcal{L}_{siam}\), further boosts performance. \(\mathcal{L}_{diac}\) improves alignment by contrasting positive and negative attributes, while \(\mathcal{L}_{siam}\) reduces intra-class variation by adjusting sensitivity to subtle differences. DAPL achieves Rank-1 accuracies of 75.20\% on CUHK-PEDES, 66.76\% on ICFG-PEDES, and 67.07\% on RSTPReid, reducing ambiguity and enhancing precision.

\textbf{Combined Effect of DTS and DAPL:}  
Combining DTS and DAPL achieves Rank-1 accuracies of 77.43\% on CUHK-PEDES, 67.87\% on ICFG-PEDES, and 69.12\% on RSTPReid, reflecting their complementary strengths—DTS for fine-grained matching and DAPL for holistic alignment

\subsection{Impact of the Number of Negative Descriptions}  
\begin{table}[htbp]
\centering
\caption{Impact of the number of negative descriptions on retrieval performance (CUHK-PEDES).}
\label{tab:negative_descriptions}
\begin{tabular}{|c|cccc|}
\hline
\textbf{Numbers} & \textbf{Rank-1} & \textbf{Rank-5} & \textbf{Rank-10} & \textbf{mAP} \\ 
\hline
0  & 77.43 & 90.73 & 94.20 & 68.30 \\
1 & 78.23 & 91.84 & 95.10 & 69.12 \\
2 & 80.05 & 92.65 & 97.48 & 70.24 \\
3 & 80.25 & 93.80 & 97.52 & 71.10 \\
4 & 79.80 & 93.45 & 96.75 & 69.80 \\
\hline
\end{tabular}
\end{table}
To evaluate the effect of varying the number of negative descriptions per image, we conducted an ablation study using our pre-trained DAPL framework on the CUHK-PEDES dataset. The study tested configurations with one, two, three, and four negative descriptions per image.
Table~\ref{tab:negative_descriptions} provides a detailed comparison of results.

The results demonstrate that using only one negative description has minimal impact compared to the baseline. When two negative descriptions are incorporated, a significant improvement is observed, particularly in Rank-1 accuracy and mAP. Adding a third negative description provides marginal gains, but the performance plateaus. Interestingly, using four negative descriptions slightly degrades performance, which we attribute to the text encoder's inability to effectively handle overly long negative descriptions. This overloading likely dilutes the importance of critical discriminative features.  

Based on these findings, we conclude that two negative descriptions strike the optimal balance between performance improvement and computational overhead. 

\subsection{Analysis of Sensitivity to Negative Descriptions}
Negative descriptions cannot independently determine exact matches. To assess DAPL’s ability to leverage negative descriptions, we augmented each query with two negative descriptions generated from the predefined attribute table. The choice of using two negative descriptions is deliberate: two selected negative descriptions effectively exclude most irrelevant candidates, while avoiding redundancy or conflicts that can arise from using more negative descriptions. As shown in Table~\ref{tab_negative}, DAPL achieves Rank-1 accuracy of 80.05\%, 71.18\%, and 70.84\% on the CUHK-PEDES, ICFG-PEDES, and RSTPReid datasets, respectively, maintaining superiority across all metrics.
These results demonstrate that DAPL enhances attribute-level detection through DTS and DIAC, while SIAM balances the influence of both positive and negative attributes, effectively utilizing negative attributes to suppress irrelevant candidates. Notably, we also observe that APTM shows competitive performance in this regard. We believe this is due to its use of an attribute-based training approach, similar to DAPL. Through our analysis of both methods’ performance, we conclude that the identification of negative attributes greatly depends on our refined focus on matching and mismatching at the attribute level.

\subsection{Comparisons on the domain generalization task}
\begin{table}[htbp]
\centering  
\caption{Cross-Domain performance (\%) between CUHK-PEDES(C) and ICFG-PEDES(I).}
\label{table:benchmark_comparison}
\begin{tabular}{|l|ccc|ccc|}
\hline
\multirow{2}{*}{Method} & \multicolumn{3}{c|}{C $\rightarrow$ I} & \multicolumn{3}{c|}{I $\rightarrow$ C} \\
& R@1 & R@5 & R@10 & R@1 & R@5 & R@10\\
\hline
SSAN~\cite{ding2021semantically}  & 29.24 & 49.00 & 58.53 &21.07 &38.94 &48.54 \\
IRRA~\cite{jiang2023cross} &41.67 &61.06  &69.24  &30.36  &52.86 &65.51\\
DCEL~\cite{li2023dcel} &43.31 &62.29 &70.31 &32.35 &54.86 &65.51 \\
\hline
DAPL &50.47 &68.62 &74.60 &45.34 &62.67 &75.43 \\
\hline
\end{tabular}
\end{table}
The retrieval performance under cross-domain settings is crucial for evaluating the applicability of text-based person search (TBPS) in real-world scenarios. Specifically, pedestrian images captured in practical environments often exhibit significant domain shifts compared to those in laboratory-collected training datasets. These domain discrepancies can lead to severe performance degradation, rendering TBPS methods ineffective in real-world applications. 

We evaluated the domain generalization capabilities of our DAPL framework on the CUHK-PEDES and ICFG-PEDES datasets. As shown in Table~\ref{table:benchmark_comparison}, DAPL significantly outperforms contemporary models such as SSAN~\cite{ding2021semantically}, IRRA~\cite{jiang2023cross}, and DCEL~\cite{li2023dcel}. Notably, DAPL achieved a Rank-1 accuracy of 50.47\% when trained on CUHK-PEDES and tested on ICFG-PEDES, surpassing its competitors. This superior performance demonstrates DAPL's robustness in handling domain variations and its ability to generalize effectively across diverse data distributions, driven by its architecture that integrates visual and textual data at a fine-grained level.

\section{Conclusion and Limitations}
In this paper, we introduce DAPL, a novel framework that enhances TBPS by integrating both negative and positive descriptions.
While effective, DAPL currently depends on a predefined attribute list for generating negative examples during training, limiting its ability to capture the full diversity of real-world scenarios. Additionally, its multi-encoder architecture, though powerful, adds complexity that may reduce efficiency in resource-constrained environments.
In summary, DAPL offers a fresh perspective on TBPS, providing a strong foundation for improving precision and adaptability in text-based person retrieval, with potential for further refinement to enhance its versatility and applicability.

\bibliographystyle{IEEEbib}
\bibliography{icme2025references}
\end{document}